%% file: main.tex
\documentclass[11pt]{article}

\PassOptionsToPackage{table}{xcolor}
\usepackage{graphicx}
\usepackage{float}

\graphicspath{
	{./images/}
	{./Figures/}
	{./Figures/Paper/}
	{./Figures/Report/}
	{../fob-results/baseline/}
	{./fob-results/baseline/}
}

\usepackage{microtype} 
\usepackage{booktabs}  
\usepackage{url}  

\usepackage{amsmath}
\usepackage{amsthm}

\usepackage{listings}
\usepackage{xcolor}
\definecolor{codegreen}{rgb}{0,0.6,0}
\definecolor{codegray}{rgb}{0.5,0.5,0.5}
\definecolor{codepurple}{rgb}{0.58,0,0.82}
\definecolor{backcolour}{rgb}{0.95,0.95,0.92}
\lstdefinestyle{mypythonstyle}{
	backgroundcolor=\color{backcolour},   
	commentstyle=\color{codegreen},
	keywordstyle=\color{magenta},
	numberstyle=\tiny\color{codegray},
	stringstyle=\color{codepurple},
	basicstyle=\ttfamily\footnotesize,
	breakatwhitespace=false,         
	breaklines=true,                 
	captionpos=b,                    
	keepspaces=true,                 
	numbers=left,                    
	numbersep=5pt,                  
	showspaces=false,                
	showstringspaces=false,
	showtabs=false,
	tabsize=4
}

\newcommand\YAMLcolonstyle{\color{black}}
\newcommand\YAMLvaluestyle{\color{blue}}
\newcommand\YAMLbracketstyle{\color{purple}}
\lstdefinelanguage{yaml}{
	keywords={true,false,null,y,n},
	keywordstyle=\color{darkgray}\bfseries,
	basicstyle=\ttfamily\footnotesize,
	sensitive=false,
	comment=[l]{\#},
	morecomment=[s]{[}{]},
	morecomment=[s]{/*}{*/},
	commentstyle=\YAMLbracketstyle,
	stringstyle=\YAMLvaluestyle\ttfamily,
	moredelim=[l][\color{orange}]{\&},
	moredelim=[l][\color{magenta}]{*},
	moredelim=**[il][\YAMLcolonstyle{:}\YAMLvaluestyle]{:},   
	morestring=[b]',
	morestring=[b]",
	tabsize=2
}

\lstdefinelanguage{yaml2}{
	keywords={true,false,null,y,n},
	keywordstyle=\color{darkgray}\bfseries,
	ndkeywords={},
	ndkeywordstyle=\color{black}\bfseries,
	identifierstyle=\color{black},
	sensitive=false,
	comment=[l]{\ },
	commentstyle=\color{blue}\ttfamily,
	stringstyle=\color{blue}\ttfamily,
	morestring=[b]',
	morestring=[b]",
	tabsize=2
}

\input{Commands/Commands.tex}

\usepackage[preprint,final]{automl}
\usepackage{cleveref}

\usepackage[%
backend=biber,
style=ext-authoryear-comp,
sortcites=true,
natbib=true,
giveninits=true,
maxcitenames=2,
doi=false,
url=true,
isbn=false,
dashed=false
]{biblatex}
\DeclareOuterCiteDelims{parencite}{\bibopenbracket}{\bibclosebracket}

\title{Fast Optimizer Benchmark}

\author[*,1]{Simon Blauth}
\author[*,1]{Tobias Bürger}
\author[*,1]{Zacharias Häringer}
\author[1]{Jörg Franke}
\author[2,1]{Frank Hutter}

\affil[*]{These authors contributed equally to this work, sharing first authorship}
\affil[1]{University of Freiburg}
\affil[2]{ELLIS Institute Tübingen}

\usepackage{hyperref}
\hypersetup{%
	pdfauthor={Simon Blauth, Tobias Bürger, Zacharias Häringer, Jörg Franke, Frank Hutter}, 
	colorlinks=true,
	linkcolor=blue,
	urlcolor=blue,
	pdftitle={Fast Optimizer Benchmark},
	pdfsubject={},
	pdfkeywords={benchmark, deep learning, neural network, training algorithms, optimizer},
}
\newcommand{\repo}[0]{\href{https://github.com/automl/FOB}{repository}}
\newcommand{\repolink}[0]{\href{https://github.com/automl/FOB}{https://github.com/automl/FOB}}

\begin{document}
	
	\maketitle
	
	\begin{abstract}
		\input{Content/abstract.tex}
	\end{abstract}
	
	
	
	
	\section{Introduction}
	\input{Content/introduction.tex}

	\section{Methods} 
	\input{Content/methods.tex}
	\section{Experiments}
	\input{Content/experiments.tex}
	\section{Broader Impact \& Limitations}
	\input{Content/statement.tex}
	\section{Conclusion}
	\input{Content/conclusion.tex}

	\begin{acknowledgements}
		The authors acknowledge support by the state of Baden-Württemberg through bwHPC and the German Research Foundation (DFG) through grant INST 35/1597-1 FUGG.
	\end{acknowledgements}
	
	
	
    \newpage
	\printbibliography
	
	
	\newpage
	\section*{Submission Checklist}
	\begin{enumerate}
		\item For all authors\dots
		\begin{enumerate}
			\item Do the main claims made in the abstract and introduction accurately
			reflect the paper's contributions and scope?
			\answerYes{}
			\item Did you describe the limitations of your work?
			\answerYes{See \cref{sec:limitation}}
			\item Did you discuss any potential negative societal impacts of your work?
			\answerYes{See \cref{sec:limitation}}
			\item Did you read the ethics review guidelines and ensure that your paper
			conforms to them? \url{https://2022.automl.cc/ethics-accessibility/}
			\answerYes{}
		\end{enumerate}
		\item If you ran experiments\dots
		\begin{enumerate}
			\item Did you use the same evaluation protocol for all methods being compared (e.g.,
			same benchmarks, data (sub)sets, available resources)?
			\answerYes{See \cref{sec:experiments}.}
			\item Did you specify all the necessary details of your evaluation (e.g., data splits,
			pre-processing, search spaces, hyperparameter tuning)?
			\answerYes{All of the experimental details are either specified in the paper or in our repository.}
			\item Did you repeat your experiments (e.g., across multiple random seeds or splits) to account for the impact of randomness in your methods or data?
			\answerYes{Some experiments were conducted on multiple seeds see \cref{sec:case-study-gridsearch}. Others had only one seed, see \cref{sec:experiments}. We always disclose the seeds used.}
			\item Did you report the uncertainty of your results (e.g., the variance across random seeds or splits)?
			\answerYes{Where applicable, see \cref{sec:case-study-gridsearch}.}
			\item Did you report the statistical significance of your results?
			\answerNo{We did not report the statistical significance explicitly. However, we mention whether results are statistically significant in our discussion, see \cref{sec:case-study-gridsearch-discussion}}.
			\item Did you use tabular or surrogate benchmarks for in-depth evaluations?
			\answerNA{Our contribution contains a tabular dataset as well as other real datasets. The evaluation was achievable using real datasets, eliminating the need for surrogate ones.}
			\item Did you compare performance over time and describe how you selected the maximum duration?
			\answerYes{We monitored the training with tensorboard and selected training durations manually. We tried to keep training as short as possible while still achieving reasonable performance.}
			\item Did you include the total amount of compute and the type of resources
			used (e.g., type of \textsc{gpu}s, internal cluster, or cloud provider)?
			\answerYes{See \cref{tab:tasks} and \cref{sec:limitation}}
			\item Did you run ablation studies to assess the impact of different
			components of your approach?
			\answerNA{Our tool offers a huge range of customization options. Running ablation studies on all of them is not feasible.}
		\end{enumerate}
		\item With respect to the code used to obtain your results\dots
		\begin{enumerate}
			\item Did you include the code, data, and instructions needed to reproduce the
			main experimental results, including all requirements (e.g.,
			\texttt{requirements.txt} with explicit versions), random seeds, an instructive
			\texttt{README} with installation, and execution commands (either in the
			supplemental material or as a \textsc{url})?
			\answerYes{See \repo.}
			\item Did you include a minimal example to replicate results on a small subset
			of the experiments or on toy data?
			\answerYes{See \repo.}
			\item Did you ensure sufficient code quality and documentation so that someone else
			can execute and understand your code?
			\answerYes{See \repo.}
			\item Did you include the raw results of running your experiments with the given
			code, data, and instructions?
			\answerYes{See release in our \repo, we added the raw results as a zip file.}
			\item Did you include the code, additional data, and instructions needed to generate
			the figures and tables in your paper based on the raw results?
			\answerYes{See \repo.}
		\end{enumerate}
		\item If you used existing assets (e.g., code, data, models)\dots
		\begin{enumerate}
			\item Did you cite the creators of used assets?
			\answerYes{}
			\item Did you discuss whether and how consent was obtained from people whose
			data you're using/curating if the license requires it?
			\answerNA{We used publicly available datasets.}
			\item Did you discuss whether the data you are using/curating contains
			personally identifiable information or offensive content?
			\answerNA{We use only publicly available datasets and existing models.}
		\end{enumerate}
		\item If you created/released new assets (e.g., code, data, models)\dots
		\begin{enumerate}
			\item Did you mention the license of the new assets (e.g., as part of your code submission)?
			\answerYes{See \repo.}
			\item Did you include the new assets either in the supplemental material or as
			a \textsc{url} (to, e.g., GitHub or Hugging Face)?
			\answerYes{See \repo.}
		\end{enumerate}
		\item If you used crowdsourcing or conducted research with human subjects\dots
		\begin{enumerate}
			\item Did you include the full text of instructions given to participants and
			screenshots, if applicable?
			\answerNA{We did not use crowdsourcing or conducted research with human subjects.}
			\item Did you describe any potential participant risks, with links to
			Institutional Review Board (\textsc{irb}) approvals, if applicable?
			\answerNA{We did not use crowdsourcing or conducted research with human subjects.}
			\item Did you include the estimated hourly wage paid to participants and the
			total amount spent on participant compensation?
			\answerNA{We did not use crowdsourcing or conducted research with human subjects.}
		\end{enumerate}
		\item If you included theoretical results\dots
		\begin{enumerate}
			\item Did you state the full set of assumptions of all theoretical results?
			\answerNA{We did not include theoretical results.}
			\item Did you include complete proofs of all theoretical results?
			\answerNA{We did not include theoretical results.}
		\end{enumerate}
	\end{enumerate}
	
	\newpage
	\appendix
	
	
	\input{Content/appendix.tex}

\end{document}

%% file: Commands/Commands.tex
\newcommand{\zz}{\mathrm{Z\kern-.3em\raise-0.5ex\hbox{Z}}}
\newcommand{\vecl}{\begin{pmatrix}}
\newcommand{\vecr}{\end{pmatrix}}

\newcommand{\veryhigh}[3]
{   \begin{tikzpicture}
	\node (tempnode-0) at (0,0) {$#1$};
	\foreach \mytext [count=\c] in {#2}
	{ \pgfmathtruncatemacro{\b}{\c-1}
		\node[above right,font=\tiny,inner sep=3pt] (tempnode-\c) at (tempnode-\b) {$\mytext$};
		\xdef\maxexp{\c}
	}
	\draw [decoration={brace,amplitude=4pt,mirror,raise=2pt},decorate] ($(tempnode-1.south east)+(-0.13,0.13)$) -- node[below right=1mm,font=\tiny] {#3} ($(tempnode-\maxexp.south east)+(-0.13,0.13)$);
	\end{tikzpicture}
}

\makeatletter
\newcommand{\dotminus}{\mathbin{\text{\@dotminus}}}
\newcommand{\@dotminus}{%
	\ooalign{\hidewidth\raise1ex\hbox{.}\hidewidth\cr$\m@th-$\cr}%
}
\makeatother

\newcommand{\swap}[2]{
	\let\temporary#1
	\let#1#2
	\let#2\temporary
	\let\temporary\undefined}



%% file: Content/abstract.tex
In this paper, we present the \textsc{Fast Optimizer Benchmark} (FOB), a tool designed for evaluating deep learning optimizers during their development. The benchmark supports tasks from multiple domains such as computer vision, natural language processing, and graph learning. The focus is on convenient usage, featuring human-readable YAML configurations, SLURM integration, and plotting utilities. FOB can be used together with existing hyperparameter optimization (HPO) tools as it handles training and resuming of runs. The modular design enables integration into custom pipelines, using it simply as a collection of tasks. 
We showcase an optimizer comparison as a usage example of our tool. FOB can be found on GitHub: \repolink.

\textbf{Keywords:} benchmark, deep learning, neural network, training algorithms, optimizer

%% file: Content/introduction.tex

Deep learning training algorithms, often simply called \emph{optimizers}, play an integral role in the training process of modern AI models.
Therefore, improving optimizers' development is a valuable contribution to the research community.
One of the challenges of developing new optimizers is to measure their performance accurately.
Researchers need fast, reliable, and reproducible performance measures starting from the early stages of the development cycle.

Our aim with this tool is to offer a user-friendly development platform that can be easily extended to meet individual needs when comparing different optimizers.
We provide tasks from different domains, baseline optimizers, and a convenient way to run experiments with our YAML configuration workflow.
Other convenience features like SLURM integration and plotting utilities are also included. 
Results can be reproduced by sharing the configuration YAML file.
A fixed collection of tasks is not only necessary for an optimizer benchmark but is also a valuable resource for other applications.
Due to the modular design, the provided tasks can be reused in custom training pipelines or together with existing HPO tools.
The tool is also convenient to set up because it does not require dataset registration, cutting down on initial setup time.

\subsection{Related Work}

In this section, we review several existing benchmarks that are relevant to our work.\\
\textbf{AlgoPerf} \parencite{algoperf} is an extensive optimizer benchmark that focuses on large tasks and implements a time-to-result benchmark with fixed hardware requirements. However, it has limitations due to the necessity for dataset registration and its restriction to large-scale tasks only.\\
\textbf{DeepOBS} \parencite{deepobs} was an early benchmark, but except for one it only features very small tasks. Additionally, it is now considered obsolete with AlgoPerf being its successor.\\
\textbf{MLPerf} \parencite{mlperf} includes a variety of large tasks but lacks a unified structure to compare optimizers independently of the tasks. Each task requires a separate setup and execution, making it less convenient for benchmarking optimizers.\\
\textbf{Deep Learning Benchmark Suite}\footnote{https://github.com/HewlettPackard/dlcookbook-dlbs} only features image datasets and focuses on evaluating accelerators and frameworks rather than optimizers.

\subsection{Measuring Optimizer Performance}

Optimizer performance can be measured in two primary ways.\\
The first method focuses on the \textit{time to achieve a certain score}, either a predefined performance threshold or the peak performance of the training run.
The second method involves \textit{fixing the training duration} and then measuring the final performance. This can be done by setting a fixed number of epochs or by limiting the wall time.\\
Another important aspect is the optimizer's sensitivity to its hyperparameters. Optimizers with good defaults that are easy to tune are more desirable, as they reduce the need for extensive tuning and make it easier to achieve optimal performance across different tasks.

The decision for FOB is to train for a fixed number of epochs and then report the best and last performance. This strategy ensures that scores are available even for underperforming optimizers or those with suboptimal hyperparameter configurations. By using epochs instead of wall time, we maintain hardware independence, although this approach can be exploited by optimizers with heavy computations. The chosen number of epochs also influences the outcome: a tight limit favors faster optimizers, while a more generous limit favors those achieving better peak performance.

%% file: Content/methods.tex

Our tool is built upon the deep learning framework PyTorch Lightning \parencite{lightning}. 
It follows the same abstraction idea and organizes essential deep learning components into exchangeable and extensible modules.
The two key modules are \textit{Tasks} and \textit{Optimizers}.
Training and evaluation are handled by the core engine.
Users can decide which task-optimizer pair to train; they can configure training parameters and optimizer hyperparameters, and specify some model parameters.
Configurations are written in human-readable YAML files.
Users can quickly assess their results with heatmaps that are being plotted after the evaluation of the model finishes.

\subsection{Usage}
Our tool tries to cover a wide range of use cases. We want users to be able to get started quickly but also give them the freedom to integrate it into their existing workflows.

\subsubsection{Running Experiments}
Users specify their experimental setup through a single YAML configuration file. Once the configuration is set, the tool handles all other aspects of running the experiment, including data loading, model training, and evaluation. This makes it easy to set up and run experiments without needing to write any code. FOB also has a built-in grid-search functionality to explore hyperparameters. Tasks and optimizers come with sensible default values, and users have the option to overwrite them as needed by simply including them in their experiment files.
We suggest referring to the corresponding default YAML files to explore all available configurable options.
The \repo{} contains several helpful examples of experiments for reference.

\subsubsection{Integration with Existing Tools}
For more advanced HPO, users can integrate FOB with existing HPO tools like SMAC \parencite{smac} or Optuna \parencite{optuna}. Although advanced HPO is not built into FOB, the tool's engine supports training and checkpointing, allowing users to resume runs. Examples using SMAC are provided in the \repo, showcasing how to implement this integration. In \cref{sec:experiments}, we present experimental results that were obtained using SMAC.

\subsubsection{Using Tasks in Your Own Pipeline}
The modular design of FOB allows researchers to use it purely as a collection of tasks and integrate them into their own frameworks or benchmarking setups. This enables researchers to benefit from FOB's task collection without being constrained by the tool’s architecture. For pointers on how you might do this see \cref{sec:python-example}.
An example of this use case is available in our repository, demonstrating how to use FOB tasks with NePS \parencite{neps}.

\subsection{Optimizer}

Besides optimizers, learning rate schedulers are important for adjusting parameter updates during training.
In this benchmark, we treat both together as the optimizer being evaluated.
A small collection of optimizers has been implemented as baselines, and to reproduce results from literature references. 
In addition to SGD with Momentum \parencite{sgd} and Adafactor \parencite{adafactor}, the baselines include implementations of AdamW \parencite{adamw} and AdamCPR \parencite{adamcpr}, all with a cosine annealing scheduler \parencite{sgdr}.

\subsubsection{Adding an Optimizer}

In an optimizer benchmark, it is a common use case to add your own optimizer and compare it against a baseline.
This process was designed to be easy and convenient. 
One can add their own implementation of an optimizer by simply following the template in the \repo{}.
The Python code is added in a single file and no further understanding of the complete code base is required.
The user has to implement a function with the following API.

\begin{lstlisting}[language=Python]
configure_optimizers(model: GroupedModel, config: OptimizerConfig)
	-> OptimizerLRScheduler:
\end{lstlisting}

\subsection{Tasks}

The benchmarking process involves multiple tasks, each comprising a predefined model architecture, training, validation, and test datasets, along with specified performance metrics.
As this benchmark aims to be budget-friendly, each task must be appropriately sized.
Training times are constrained to ensure completion of the full suite within one day on a single node with 4 GPUs.
The tasks cover a range of domains, model architectures, and sizes to ensure generalization and fair comparisons.
Popular domains, such as computer vision and natural language processing, have been allocated the largest share of training time.

The \textit{MNIST, Classification Small, Classification} and \textit{Segmentation} tasks from the \textbf{computer vision} domain utilize the MNIST, CIFAR-100, ImageNet-64, and ADE20K datasets \parencite{mnist, cifar100, imagenet64, ade20k}. We employ models such as MLP, ResNet, Wide-ResNet, and SegFormer \parencite{resnet, wide-resnet, segformer}.

In the \textit{Translation} task from the \textbf{natural language processing} domain, the goal is to translate between English and German sentences from the WMT17 dataset \parencite{bojar-graham-kamran:2017:WMT} using a small T5 transformer \parencite{t5-transformer}.

For the \textit{Graph Tiny} and \textit{Graph} tasks, we utilize the Cora dataset and the molecular ogb-molhiv dataset \parencite{McCallum2000, ogb-molhiv}.
We leverage \textbf{graph} neural network models such as a simple Graph Convolutional Network (GCN) and a Graph Isomorphism Network (GIN) \parencite{gcn, gin}.

Lastly the \textit{Tabular} task employs a FT-Transformer on the \textbf{tabular} California Housing dataset \parencite{ft-transformer, KELLEYPACE1997291}.

\begin{table}[h]
	\centering
	\begin{tabular}{|l|c|c|c|c|}
		\hline
		Dataset &
		Model &
		Problem Type &
		GPUs &
		Runtime \\
		\hline\hline
		
		MNIST &
		MLP &
		Image Classification &
		1 &
		1~min  \\
		\hline
		Cora &
		GCN &
		Node Classification &
		1 &
		1~min \\
		\hline
		California Housing &
		FT Transformer &
		Tabular Regression &
		1 &
		2~min \\
		\hline
		CIFAR100 &
		Resnet18 &
		Image Classification &
		1 &
		10~min  \\
		\hline
		ogbg-molhiv &
		GIN &
		Graph Property Prediction &
		1 &
		20~min  \\
		\hline
		Imagenet-64x64 &
		Wide ResNet &
		Image Classification &
		4 &
		4~h \\
		\hline
		MIT Scene Parse &
		SegFormer &
		Semantic Segmentation &
		4 &
		5~h \\
		\hline
		WMT17(en-de) &
		T5 small &
		Machine Translation &
		4 &
		6~h \\
		\hline
	\end{tabular}
	\caption{The ensemble of tasks in FOB. Runtime was measured on 4xA100 40GB GPUs, 128GB RAM. For comprehensive details, refer to the \repo{}.}
	\label{tab:tasks}
\end{table}

\subsubsection{Modifying a Task}

Minor modifications can be applied to the models used, with certain parameters already accessible through the YAML configuration files.
Common examples include adjustments to train transformations, dropout rates, and hidden channel sizes.
Furthermore, users can easily adapt existing tasks by making a single code change and adding an entry to the \texttt{default.yaml} file, such as modifying the activation function.

\noindent
Users can also introduce their own tasks, enabling coverage of more domains where specific optimizers may excel.
To facilitate user contributions, the \repo{} includes a template and instructions for adding custom tasks.

%% file: Content/experiments.tex
\label{sec:experiments}
We conducted experiments using SMAC \parencite{smac} on four of our smaller tasks to evaluate the performance of three optimizers: AdamW, AdamCPR, and SGD \parencite{adamw, adamcpr, sgd}.
For experiments utilizing our full suite of tasks, refer to \cref{sec:case-study-gridsearch}.

The hyperparameters of each optimizer were tuned using SMAC, see \cref{sec:supp} for the search space.
Multi-Fidelity-Optimization was used with the training epochs as budget.
The intensifier used was Hyperband \parencite{hyperband} with $\eta=3$.
Each optimizer was tuned for 250 trials with 10\% initial configurations.
For each optimizer, we took the best configuration returned by SMAC and retrained them on three seeds $s \in \{1,2,3\}$ to obtain the test performance detailed in \cref{tab:smac-performance}.
For each task and optimizer, we report the test performance of the last model checkpoint as well as the test performance of the model checkpoint with the best validation performance.
The models were trained using A100 and A40 GPUs on the HELIX cluster.

\begin{table}[h]
	\centering
	\begin{tabular}{|l|c|c|c|c|}
		\hline
		Task & Class. Small & Graph & Graph Tiny & Tabular \\ \hline
        Metric    & Accuracy $\uparrow$ & ROC AUC $\uparrow$ & Accuracy $\uparrow$ & RMSE $\downarrow$ \\ \hline \hline
		AdamCPR best & $77.2 \pm 0.16$ & $75.57 \pm 3.13$ & $80.27 \pm 0.32$ & $0.403 \pm 0.011$ \\ \hline
		AdamW best & $\mathbf{77.29 \pm 0.17}$ \cellcolor{lightgray} & $75.75 \pm 1.04$ & $80.63 \pm 1.0$ & $0.401 \pm 0.008$ \\ \hline
		SGD best & $77.27 \pm 0.28$ & $75.89 \pm 0.58$ & $14.47 \pm 3.45$ & $0.419 \pm 0.007$ \\ \hline
		AdamCPR last & $77.1 \pm 0.2$ & $\mathbf{76.86 \pm 1.06}$ \cellcolor{lightgray} & $79.5 \pm 0.61$ & $\mathbf{0.399 \pm 0.005} \cellcolor{lightgray} $ \\ \hline
		AdamW last & $77.18 \pm 0.14$ & $75.86 \pm 0.35$ & $\mathbf{81.4 \pm 0.35} \cellcolor{lightgray} $ & $0.401 \pm 0.005$ \\ \hline
		SGD last & $77.0 \pm 0.49$ & $75.99 \pm 1.34$ & $14.83 \pm 3.67$ & $0.421 \pm 0.008$ \\ \hline
	\end{tabular}
    
	\caption{Performance per task of best hyperparameter configuration found by SMAC evaluated on three seeds.}
	\label{tab:smac-performance}
\end{table}


\subsection{Discussion}
On the \textit{Classification Small} task, all optimizers performed similarly, with AdamW slightly outperforming the others. For the \textit{Graph} task, AdamCPR last achieved the highest performance, demonstrating an advantage over the other optimizers. On the \textit{Graph Tiny} task, AdamW achieved the highest score, while SGD performed poorly. In the \textit{Tabular} task, the results were similar across all optimizers, with AdamCPR achieving the best result.

Overall, both AdamW and AdamCPR performed similarly, obtaining consistent results across the tasks. In contrast, SGD exhibited clear weaknesses, particularly in the \textit{Graph Tiny} and \textit{Tabular} tasks, where its performance was significantly lower than that of AdamW and AdamCPR.

%% file: Content/statement.tex
%
%
%
%
%
\label{sec:limitation}
Large-scale benchmarks can require high computational efforts. We aim to mitigate this issue by including small tasks in our benchmark suite. When running the entire suite, especially with many hyperparameter configurations, the computations can still be expensive. For example our experiment detailed in \cref{sec:case-study-gridsearch} required approximately 2400 GPU hours on A100 GPUs. With an estimated carbon efficiency of 0.432 kg of CO2 per kWh, the total carbon emission from this experiment is approximately 260 kg of CO2. In contrast, our experiment in \cref{sec:experiments}, which only used smaller tasks, required approximately 165 GPU hours and 18 kg of CO2. This shows that the inclusion of smaller tasks with a modular approach enables researchers to conduct experiments with smaller environmental footprints. 
We do not expect any new societal risk to arise from this work since we only use already existing models and publicly available datasets.

The limitations of our contribution include the absence of very large tasks in our benchmark, which might lead to some aspects not being captured correctly. However, there are already tools that achieve this (e.g. AlgoPerf \parencite{algoperf}). Also, some fields of deep learning like speech recognition are not covered. Since our tool is designed to be extensible, we hope that these gaps might be filled in the future. Since we built our tool upon PyTorch Lightning \parencite{lightning}, we are restricted to the hardware supported by that framework, so some hardware might not work with FOB.

%% file: Content/conclusion.tex
In this paper, we introduced the \textbf{Fast Optimizer Benchmark (FOB)}, a convenient and extensible tool designed for evaluating deep learning optimizers across diverse tasks. FOB supports tasks from multiple domains, offers easy configuration through YAML files, and includes features such as SLURM integration to support distributed computing on clusters. Additionally, the modular design of FOB facilitates its use for various purposes, including hyperparameter optimization (HPO), by allowing easy integration with other tools. We showcased the usage of FOB and conducted experiments to demonstrate its capabilities.

An important open question remains for future work: To what degree do the results of this benchmark correlate with larger, related benchmarks? Establishing a reliable and cost-effective proxy for extensive benchmarks like AlgoPerf would greatly benefit the research community.

%% file: Content/appendix.tex
\section{Hyperparameter Search Spaces}
\label{sec:supp}

The searchspace detailed in \cref{tab:hyperparameters} was used as an input for SMAC \parencite{smac}.
For more details on the experiment refer to \cref{sec:experiments}.

\begin{table}[h]
\centering
\begin{tabular}{|l|c|c|c|}
	\hline
	Hyperparameter & AdamW  & AdamCPR & SGD \\ \hline \hline
	Learning Rate  & Log $[1\text{e-}5, 1\text{e-}1]$ & Log $[1\text{e-}5, 1\text{e-}1]$ & Log $[1\text{e-}5, 1\text{e-}1]$ \\ \hline
	Minimal LR & Log $[0.1\%, 10\%]$ & Log $[0.1\%, 10\%]$ & Log $[0.1\%, 10\%]$ \\ \hline
	LR Warmup & Log $[0.1\%, 100\%]$ & Log $[0.1\%, 100\%]$ & Log $[0.1\%, 100\%]$ \\ \hline
	Weight Decay & Log $[1\text{e-}5, 1]$ & -                          & Log $[1\text{e-}5, 1]$ \\ \hline
	$1 - \beta_1$ & Log $[1\text{e-}2, 2\text{e-}1]$ & Log $[1\text{e-}2, 2\text{e-}1]$ & -                          \\ \hline
	$\beta_2$ & $[0.9, 0.999]$          & $[0.9, 0.999]$ & - \\ \hline
	Momentum & - & - & $[0, 1]$ \\ \hline
	Kappa Init Param & -  & Log $[1, 19550]$  & - \\ \hline
	Kappa Init Method & - & warm\_start       & - \\ \hline
\end{tabular}
\caption{The respective hyperparameter search spaces for the AdamW, AdamCPR and SGD optimizers used in the SMAC experiment.}
\label{tab:hyperparameters}
\end{table}

\section{Usage Example: YAML configuration}
\label{sec:yaml-config}

Here we provide some examples of how a user can configure FOB using YAML configuration files.

\begin{figure}[h]
	\begin{minipage}{0.45\textwidth}
\begin{lstlisting}[language=yaml]
task:
	name: mnist
	max_epochs: 10
	model:
		num_hidden: 42
optimizer:
	name: adamw_baseline
	learning_rate: 1.0e-2
\end{lstlisting}
	\end{minipage}
	\hfill
	\begin{minipage}{0.45\textwidth}
		Training on the \textit{MNIST} task for 10 epochs.
		It is possible to make some changes to the model like the number of hidden units.
		AdamW is used as an optimizer and the learning rate is set.
		All parameters that are not specified in the configuration are taken from a default file.
	\end{minipage}
	\label{fig:yaml-file-1}
	\caption{A YAML configuration to start a small MNIST experiment.}
\end{figure}

\begin{figure}[h]
	\begin{minipage}{0.45\textwidth}
\begin{lstlisting}[language=yaml]
task:
	name: mnist
	max_epochs: 10
	model:
		num_hidden: [128, 256]
optimizer:
	- name: adamw_baseline
		beta2: 0.98
	- name: sgd_baseline
		momentum: 0.5
engine:
	seed: [42, 47]
\end{lstlisting}
\end{minipage}
\hfill
\begin{minipage}{0.45\textwidth}
	This example showcases how we can specify lists in the YAML file to perform a gridsearch.
	FOB essentially computes the cartesian product of all lists in an experiment configuration.
	In this example we have 2 model sizes, 2 optimizers and 2 seeds, resulting in 8 training runs in total.
\end{minipage}
\label{fig:yaml-file-2}
\caption{A YAML configuration to compare different settings on the MNIST task.}
\end{figure}

\begin{figure}[H]
	\begin{minipage}{0.45\textwidth}
\begin{lstlisting}[language=yaml]
task:
	name: classification_small
optimizer:
	- name: adamcpr_fast
		learning_rate:
			[1.e-1, 1.e-2, 1.e-3, 1.e-4]
		kappa_init_param:
			[1, 2, 4, 8, 16]
	- name: adamw_baseline
		learning_rate:
			[1.e-1, 1.e-2, 1.e-3, 1.e-4]
		weight_decay:
			[10, 1, 1.e-1, 1.e-2, 1.e-3]
engine:
	seed: [1, 2, 3]
evaluation:
	output_types: [pdf, png]
	plot:
		x_axis:
			- optimizer.kappa_init_param
			- optimizer.weight_decay
\end{lstlisting}
\end{minipage}
\hfill
\begin{minipage}{0.45\textwidth}
This is the configuration used to perform the experiment of \cref{fig:classification-small-best}.
Here we have 2 optimizers each evaluated for 4x5 hyperparameter configurations on 3 seeds each for a total of 120 trials.
We also showcase our plotting utilities which are specified in the same YAML file.
Running this file directly results in \cref{fig:classification-small-best}.
\end{minipage}
\label{fig:yaml-file-3}
\caption{The YAML configuration used to create the comparison on \textit{Classification Small} in \cref{fig:classification-small-best}.}
\end{figure}

\section{Python example}
\label{sec:python-example}
Below we give you a starting point if you want to handle training yourself but still want to make use of the tasks from FOB. This snippet shows that given a configuration (like the result from parsing a YAML file) you can easily access the Lightning Module and Datamodule comprising the task.
\begin{figure}[H]
\begin{lstlisting}[language=python]
from pytorch_fob import Engine
# create your config, e.g.
config = {"task": "mnist"}
engine = Engine()
engine.parse_experiment(config)
run = next(engine.runs())
model, datamodule = run.get_task()
# run your custom training pipeline...
\end{lstlisting}
\caption{Using FOB tasks in your own pipeline.}
\end{figure}

\section{Case Study: comparing AdamCPR and AdamW}
\label{sec:case-study-gridsearch}
To demonstrate the benefits of our contribution, we include a case study of comparing the \textbf{AdamW} \parencite{adamw} and the \textbf{AdamCPR} \parencite{adamcpr} optimizer on our set of tasks.

\subsection{Experimental Setup}
\label{sec:experimental-setup}
We use a Cosine-Annealing learning rate scheduler with linear warmup.
The minimum learning rate of the cosine decay is set to $1\%$ of the initial learning rate.
For all tasks, we perform a gridsearch over two hyperparameters.
Both optimizers sweep over the learning rate, while the second hyperparameter depends on the optimizer: weight decay for AdamW and the Kappa-Init-Parameter for AdamCPR.
We choose the 'warm-start' ($\verb*|Kappa-I|_{s}$) setting for AdamCPR, so the Kappa-Init-Parameter corresponds to the number of steps until fixing the regularization.
We implemented the Kappa-Init-Parameter as a factor of the steps for the learning rate warmup.
For the number of learning rate warmup steps, we use $1\%$ of the total steps.
This means that the Kappa-Init-Parameter can be a value between 0 and 100.

\noindent
For the choice of values in the search grid we follow \parencite[Section~5.2]{adamcpr} and use values evenly spaced on a $\log_{10}$ scale for learning rate and weight decay, while the values for the Kappa-Init-Parameter are evenly spaced on a $\log_{2}$ scale.
We usually started with a small grid and expanded it in the directions where the best values were at the borders.

\noindent
The models were trained using A100 and A40 GPUs on the HELIX cluster. For the exact details please refer to the baseline configurations in the \repo{}.

\subsection{Results}
Here we present the results of our experiments grouped by each task.
All experiments were repeated with three seeds $s \in \{1,2,3\}$, over which we computed the standard deviation.
For each task and optimizer, we report the test performance of the last model checkpoint as well as the test performance of the model checkpoint with the best validation performance.

\begin{table}
    \centering
    \begin{tabular}{|c|c|c|c|c|} 
    \hline
    Task         & Classification                                                & Class. Small                                                  & MNIST                                                         & Segmentation                                                   \\ 
    \hline
    Metric       & Accuracy $\uparrow$                                           & Accuracy $\uparrow$                                           & Accuracy $\uparrow$                                           & mIoU $\uparrow$                                                \\ 
    \hline \hline
    AdamCPR best & $69.15 \pm 0.07$                                              & $77.54 \pm 0.16$                                              & $97.29 \pm 0.19$                                              & \cellcolor{lightgray}$\mathbf{35.59 \pm 0.21}$  \\ 
    \hline
    AdamW best   & \cellcolor{lightgray}$\mathbf{69.79 \pm 0.04}$ & \cellcolor{lightgray}$\mathbf{77.87 \pm 0.25}$ & $97.61 \pm 0.08$                                              & $35.52 \pm 0.12$                                               \\ 
    \hline
    AdamCPR last & $69.22 \pm 0.08$                                              & $77.54 \pm 0.16$                                              & $97.34 \pm 0.19$                                              & $35.35 \pm 0.24$                                               \\ 
    \hline
    AdamW last   & $69.67 \pm 0.1$                                               & $77.79 \pm 0.25$                                              & \cellcolor{lightgray}$\mathbf{97.63 \pm 0.08}$ & $35.45 \pm 0.19$                                               \\
    \hline
    \end{tabular}
    \label{tab:gridsearch-performance-cv}
    \caption{Maximal performance of the computer vision tasks gridsearch. All values are given in \%.}
\end{table}

\begin{table}
    \centering
    \begin{tabular}{|c|c|c|c|c|} 
    \hline
    Task         & Graph                                                        & Graph Tiny                                                    & Tabular                                                        & Translation                                                  \\ 
    \hline
    Metric       & ROC AUC $\uparrow$                                           & Accuracy $\uparrow$                                           & RMSE $\downarrow$                                              & BLEU $\uparrow$                                              \\ 
    \hline \hline
    AdamCPR best & $76.72 \pm 2.12$                                             & $81.93 \pm 0.72$                                              & \cellcolor{lightgray}$\mathbf{0.396 \pm 0.004}$ & \cellcolor{lightgray}$\mathbf{26.4 \pm 0.1}$  \\ 
    \hline
    AdamW best   & $77.18 \pm 1.4$                                              & \cellcolor{lightgray}$\mathbf{82.03 \pm 0.64}$ & $0.398 \pm 0.006$                                              & $26.25 \pm 0.22$                                             \\ 
    \hline
    AdamCPR last & $76.42 \pm 0.45$                                             & $79.9 \pm 0.7$                                                & $0.398 \pm 0.005$                                              & $26.4 \pm 0.1$                                               \\ 
    \hline
    AdamW last   & \cellcolor{lightgray}$\mathbf{77.4 \pm 1.31}$ & $80.1 \pm 0.82$                                               & $0.397 \pm 0.005$                                              & $26.25 \pm 0.22$                                             \\
    \hline
    \end{tabular}
    \label{tab:gridsearch-performance-non-cv}
    \caption{Maximal performance of the gridsearch on the \textit{Graph}, \textit{Graph Tiny}, \textit{Tabular} and \textit{Translation} tasks. All values except for RMSE are given in \%.}
\end{table}

\noindent
For each task, we present plots of the performance across the entire search grid.
The plots show the test performance of the checkpoint with the highest validation performance, essentially an early-stopping setting.

\begin{figure}[H]
	\includegraphics[width=.95\textwidth]{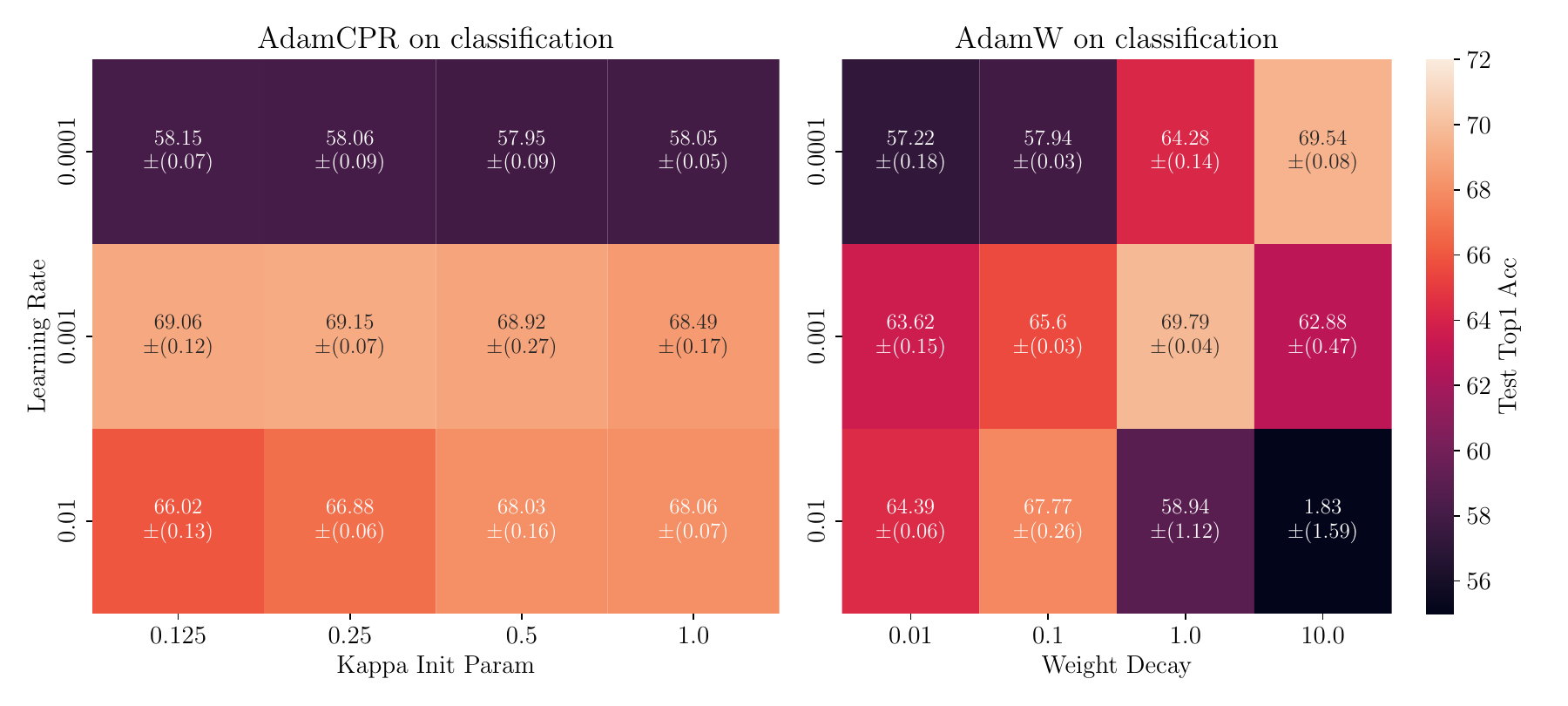}
	\caption{AdamW needs high weight decay regularization. \\AdamCPR performs best with few warmup steps.}
	\label{fig:classification-best}
\end{figure}

\begin{figure}[H]
	\includegraphics[width=.95\textwidth]{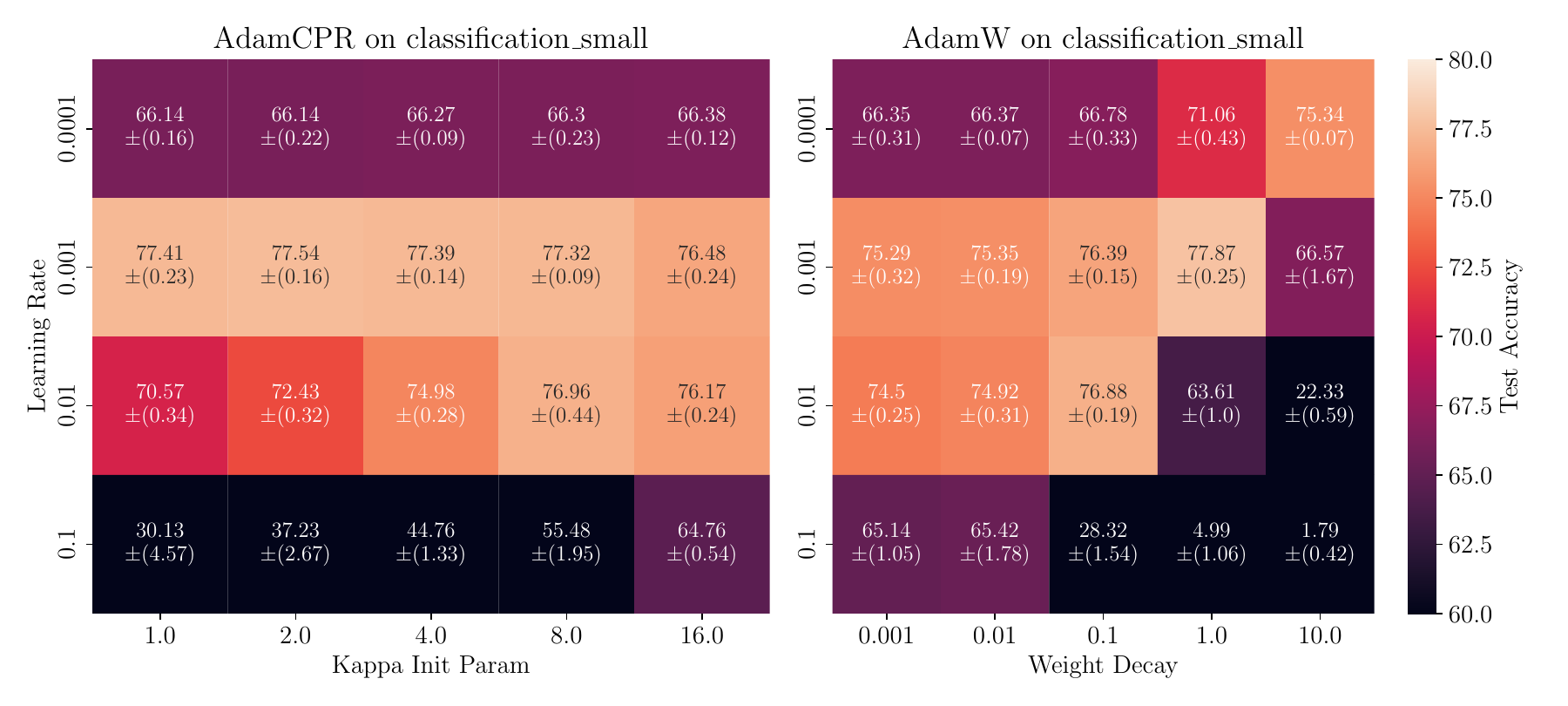}
	\caption{
		The best hyperparameters for AdamW are the same as for the \textit{Classification} task (\cref{fig:classification-best}).
		Generally, the grid elements that overlap between these tasks look similar in terms of relative performance.
		For AdamCPR we sadly cannot make this comparison as we chose different values between the tasks.
		However, we can observe that while peak performance is better for AdamW, more configurations reach a high score for AdamCPR.\\
		Note that our hyperparameters are different from the ones used in \parencite{adamcpr}. For a comparison more closely following the original, check \cref{sec:appendix-a}.
	}
	\label{fig:classification-small-best}
\end{figure}

\begin{figure}[H]
	\includegraphics[width=.95\textwidth]{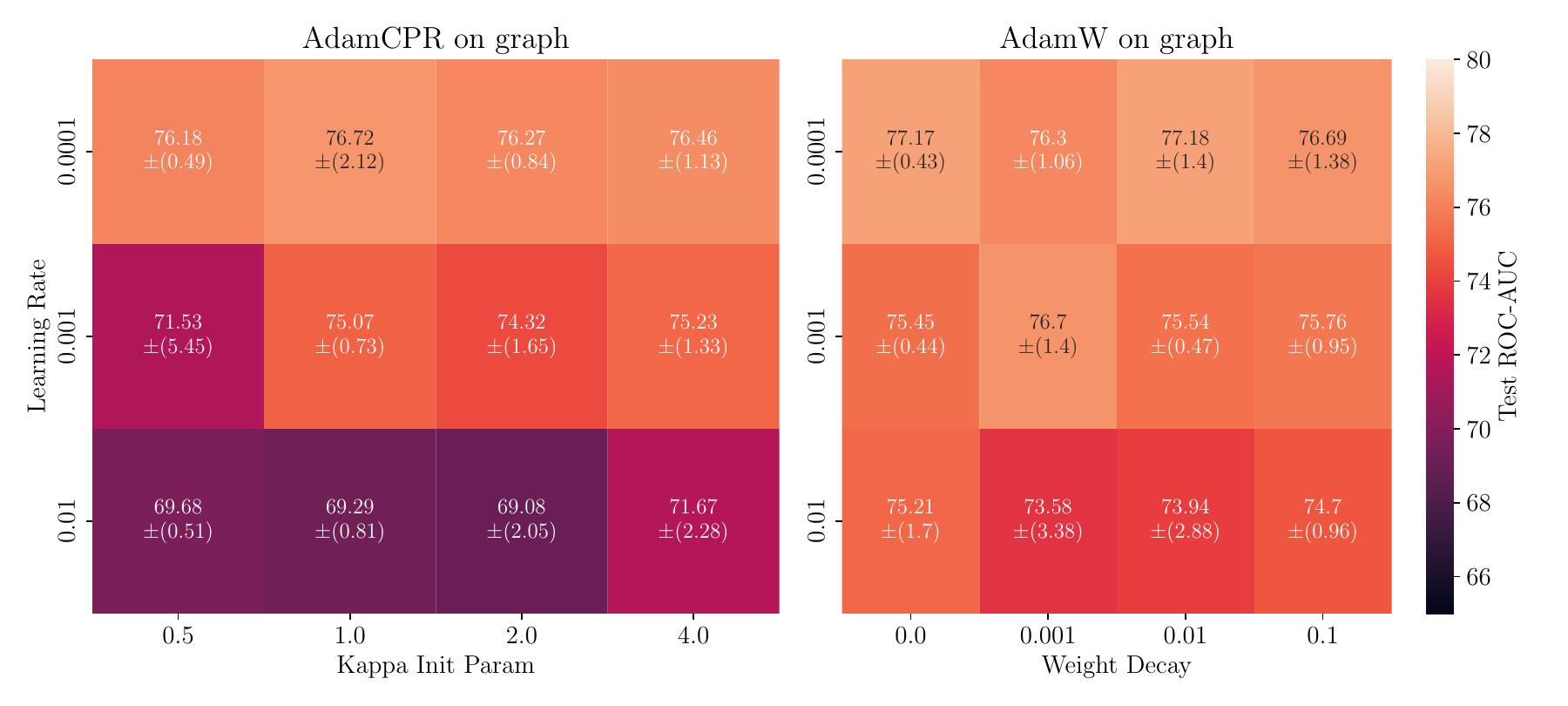}
	\caption{AdamW outperforms AdamCPR on the \textit{Graph} task.}
	\label{fig:graph-best}
\end{figure}

\begin{figure}[H]
	\includegraphics[width=.95\textwidth]{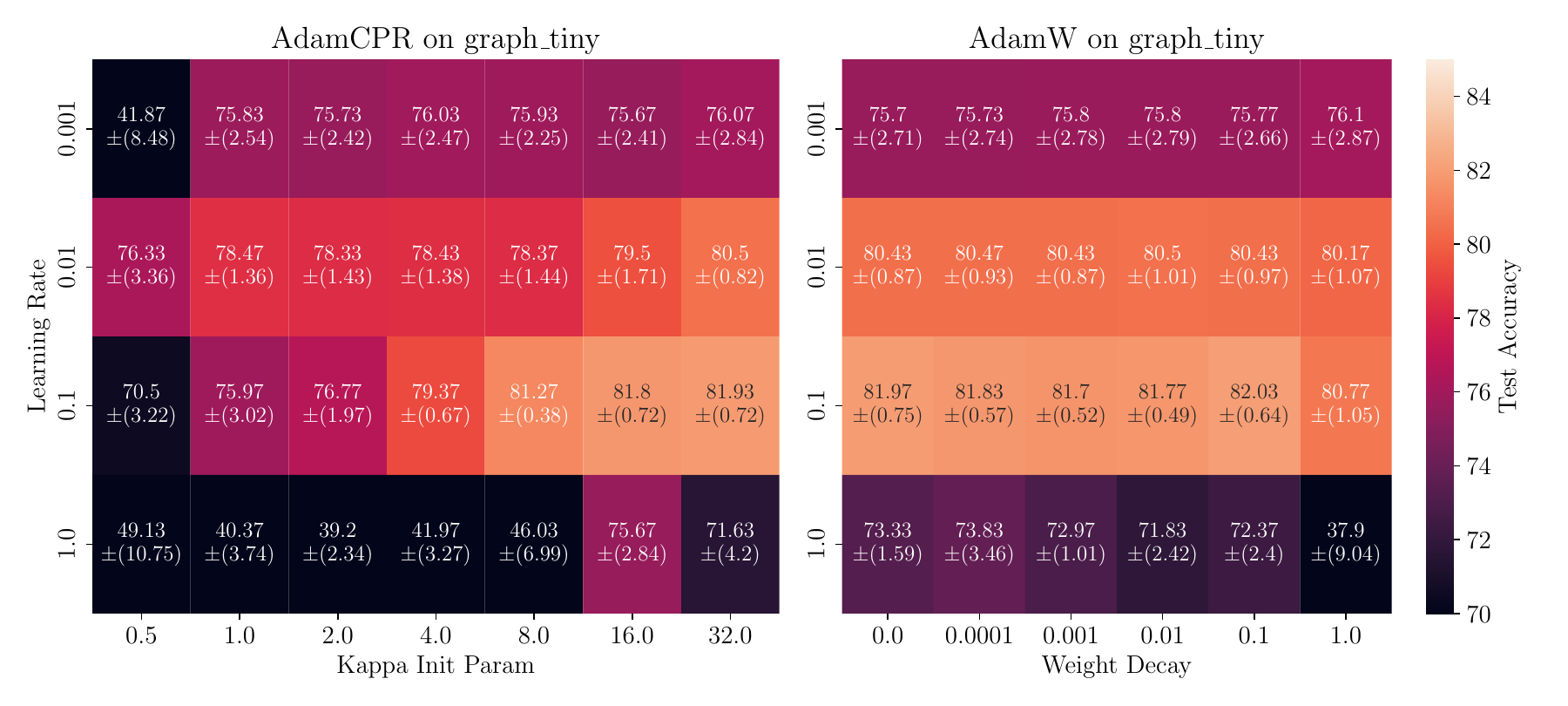}
	\caption{On \textit{Graph Tiny} there is little to no response to weight decay. For AdamW there is almost no change in performance for different weight decay values. AdamCPR performs poorly for this task and converges to AdamW without weight decay for higher Kappa-Init-Parameter (CPR warmup steps).}
	\label{fig:graph-tiny-best}
\end{figure}

\begin{figure}[H]
	\includegraphics[width=.95\textwidth]{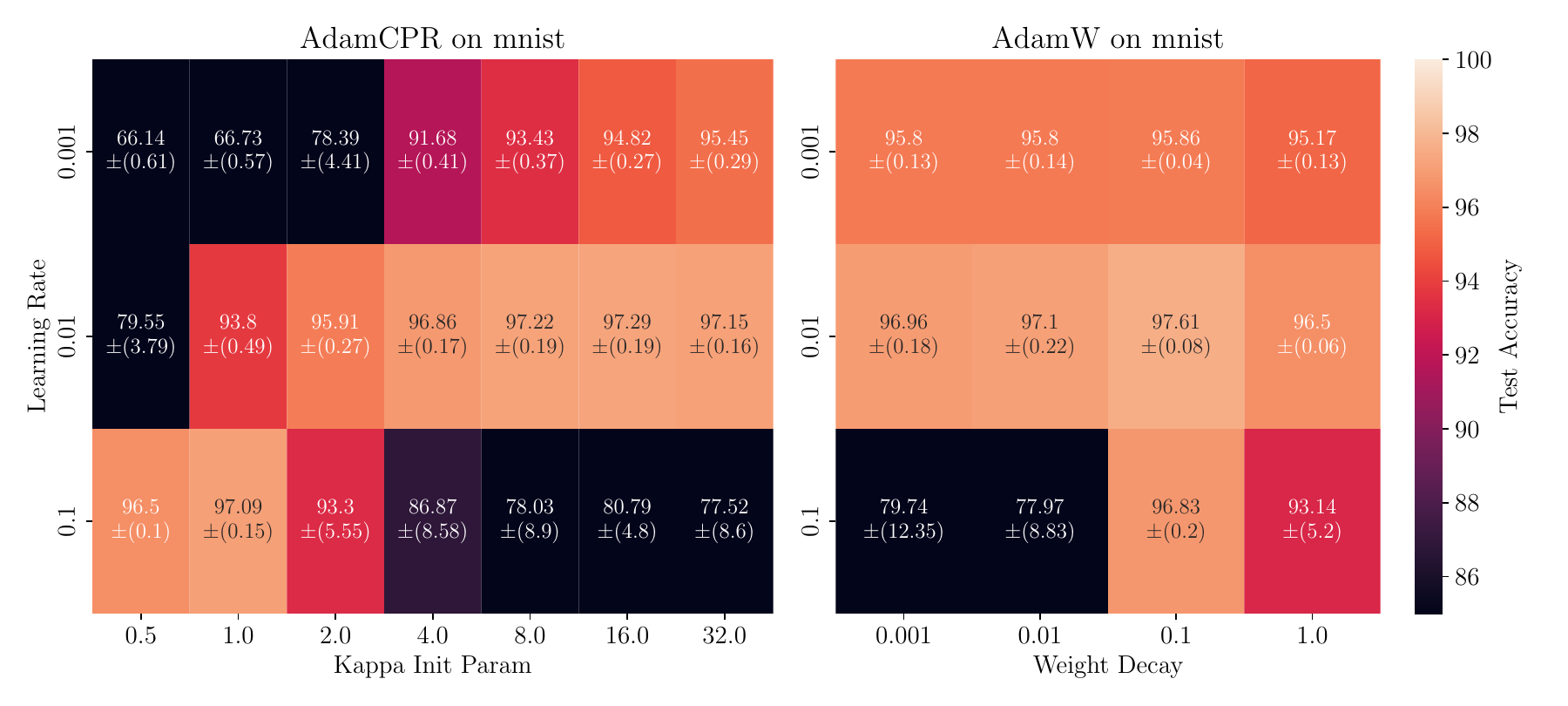}
	\caption{The performance on \textit{MNIST} behaves similar to the one on \textit{Graph Tiny} (\ref{fig:graph-tiny-best}); AdamCPR seems to offer little benefit over plain AdamW without weight decay.}
	\label{fig:mnist-best}
\end{figure}

\begin{figure}[H]
	\includegraphics[width=.95\textwidth]{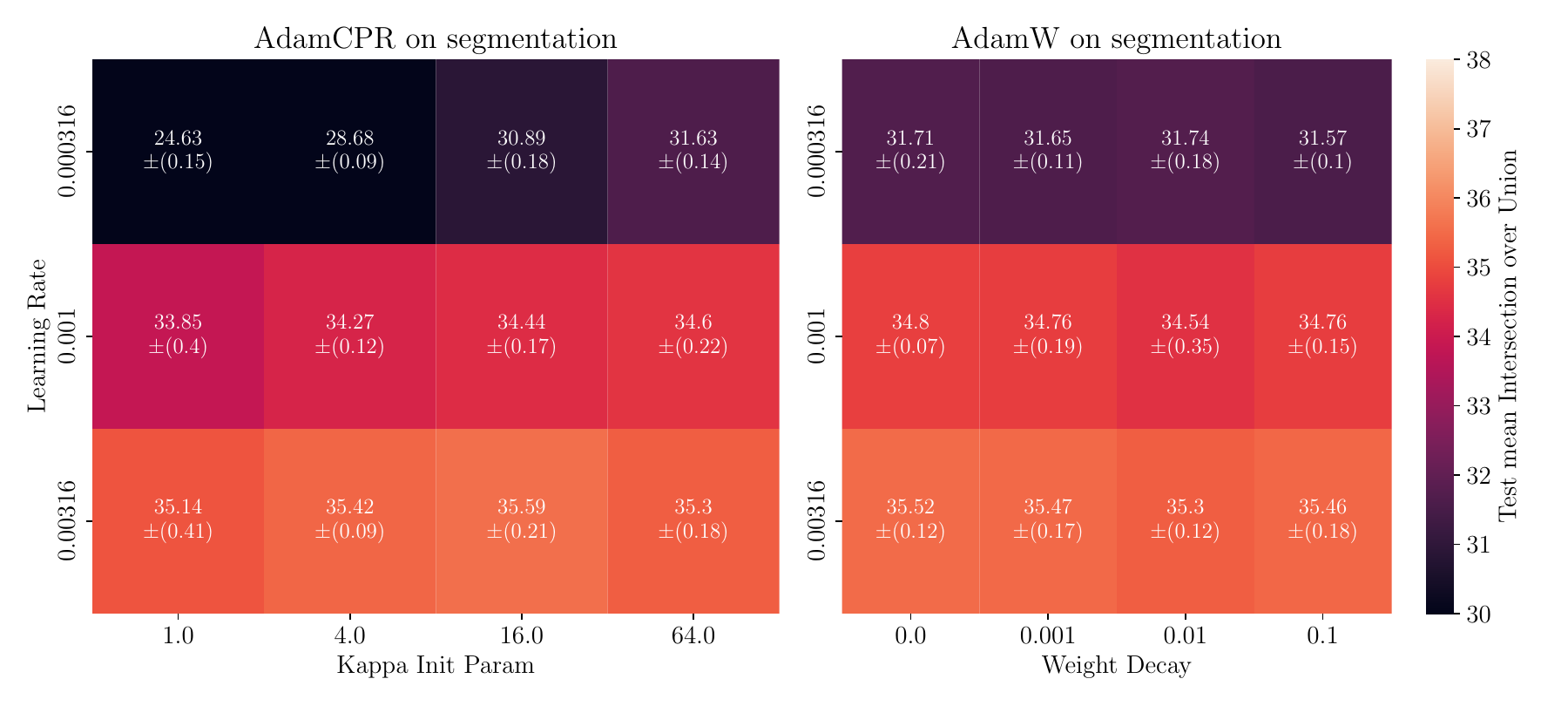}
	\caption{\textit{Segmentation} shows minimal sensitivity to weight decay adjustments and changes in the number of CPR warmup steps. However, AdamCPR performs slightly better here.}
	\label{fig:segmentation-best}
\end{figure}

\begin{figure}[H]
	\includegraphics[width=.95\textwidth]{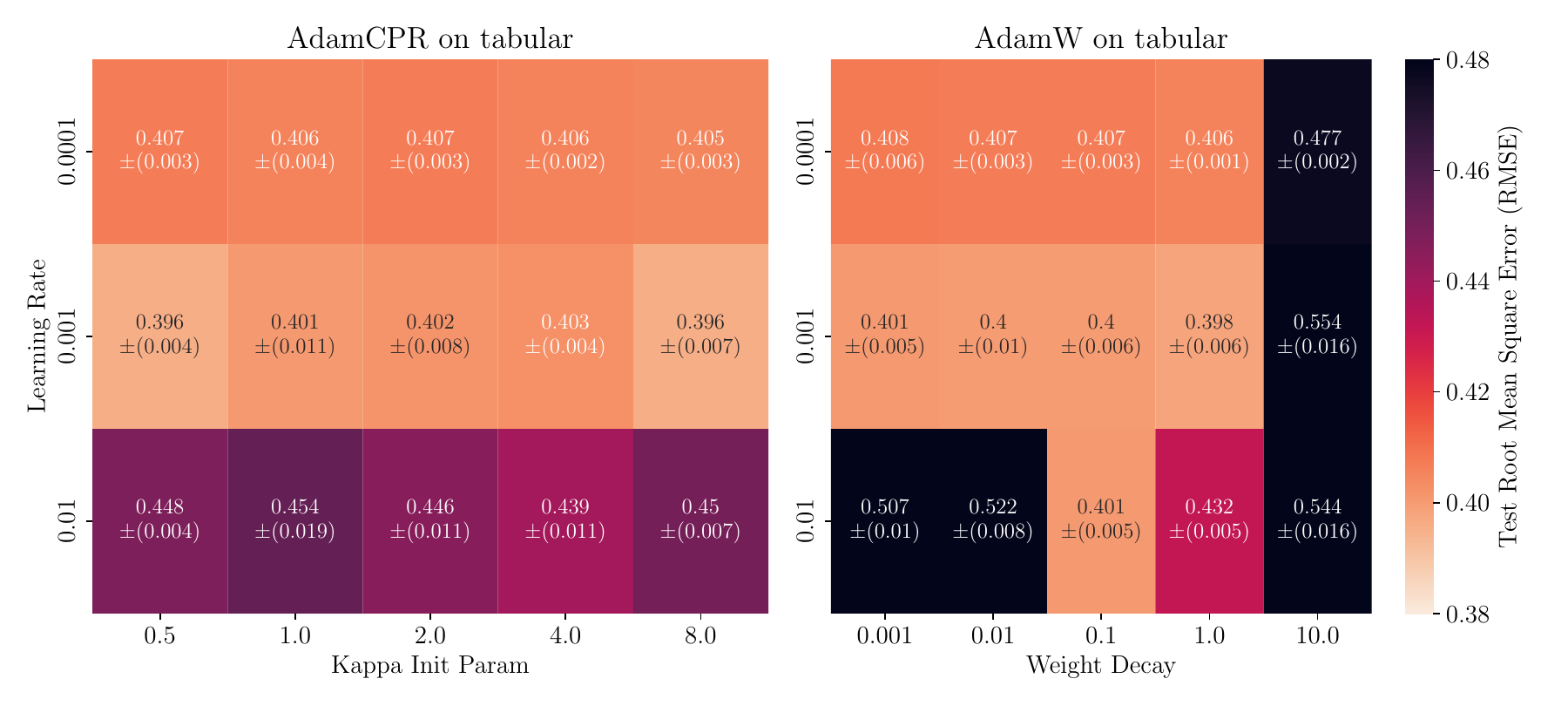}
	\caption{Both optimizers perform well on the \texttt{tabular} task, nevertheless AdamCPR performs slightly better.}
	\label{fig:tabular-best}
\end{figure}

\begin{figure}[H]
	\includegraphics[width=.95\textwidth]{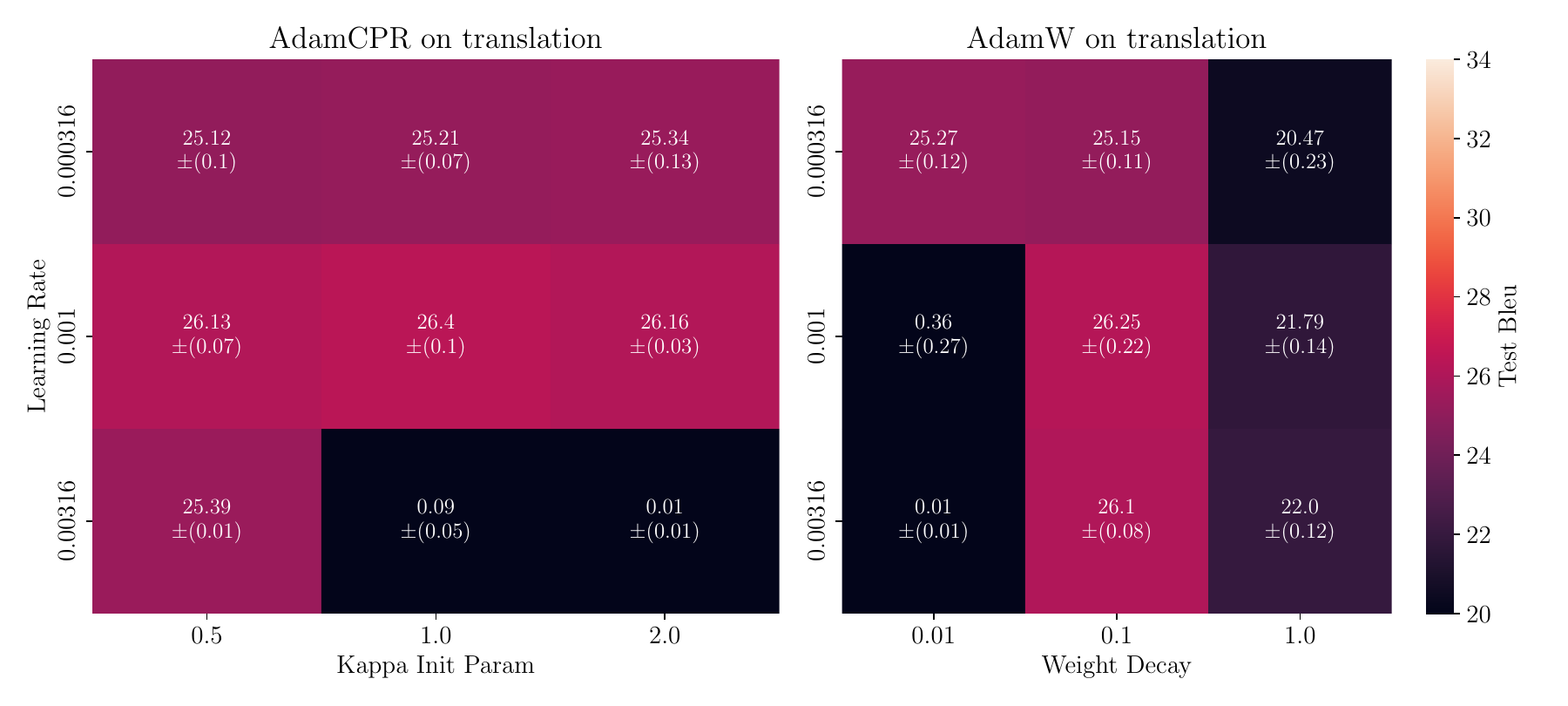}
	\caption{On the \textit{Translation} task AdamCPR outperforms AdamW. However, AdamW shows high responsiveness to weight decay. A better configuration might be found when searching closer to the current optimal weight decay.}
	\label{fig:translation-best}
\end{figure}

\subsection{Discussion}
\label{sec:case-study-gridsearch-discussion}

When comparing the results of these experiments, one should consider their purpose.
They were not solely conducted to compare the two optimizers but were also driven by the goal of exploring the best achievable performance, to set appropriate standards.

\noindent
In our experiments, the quality of optimizers is primarily evaluated by their peak performance.
However, only a limited number of configurations are explored and it is unlikely that a global optimum was found. 
We assume that the influence of convergence speed is not substantial since the highest validation performance was usually before the final epoch.
Reasonable initial search spaces were iteratively expanded to find the best configurations.
Thus the sensitivity of the optimizer to its hyperparameters had some influence on the measured peak performance.
In the sense that a robust optimizer has a higher chance of finding a good hyperparameter configuration.
Since both optimizers heavily depend on their hyperparameters, it was not feasible to use the same values across all tasks without significantly increasing the experiment budget or sacrificing peak performance.

\noindent
In our experiments, both optimizers perform similarly.
It depends on the task which optimizer has the better score.
However, the difference is usually really small, to a point where it probably is negligible and shows little statistical significance.
While either shows its own characteristics, we can not determine a superior optimizer based on these results.


\subsection{AdamCPR on Classification Small}
\label{sec:appendix-a}
Here we present a setup close to the experiment in \parencite[Section~5.2]{adamcpr}.
The hyperparamters were set to match those of the authors and we use a subset of their search grid.
Most notably, we set the minimum learning rate to 10\% its initial value instead of 1\% its initial value as in our experiment (\cref{sec:experimental-setup}).
Also we do not use trivial augment and add \verb*|label_smoothing=0.1|.
The number of learning rate warmup steps is set to 500.

\begin{figure}[H]
	\includegraphics[width=.95\textwidth]{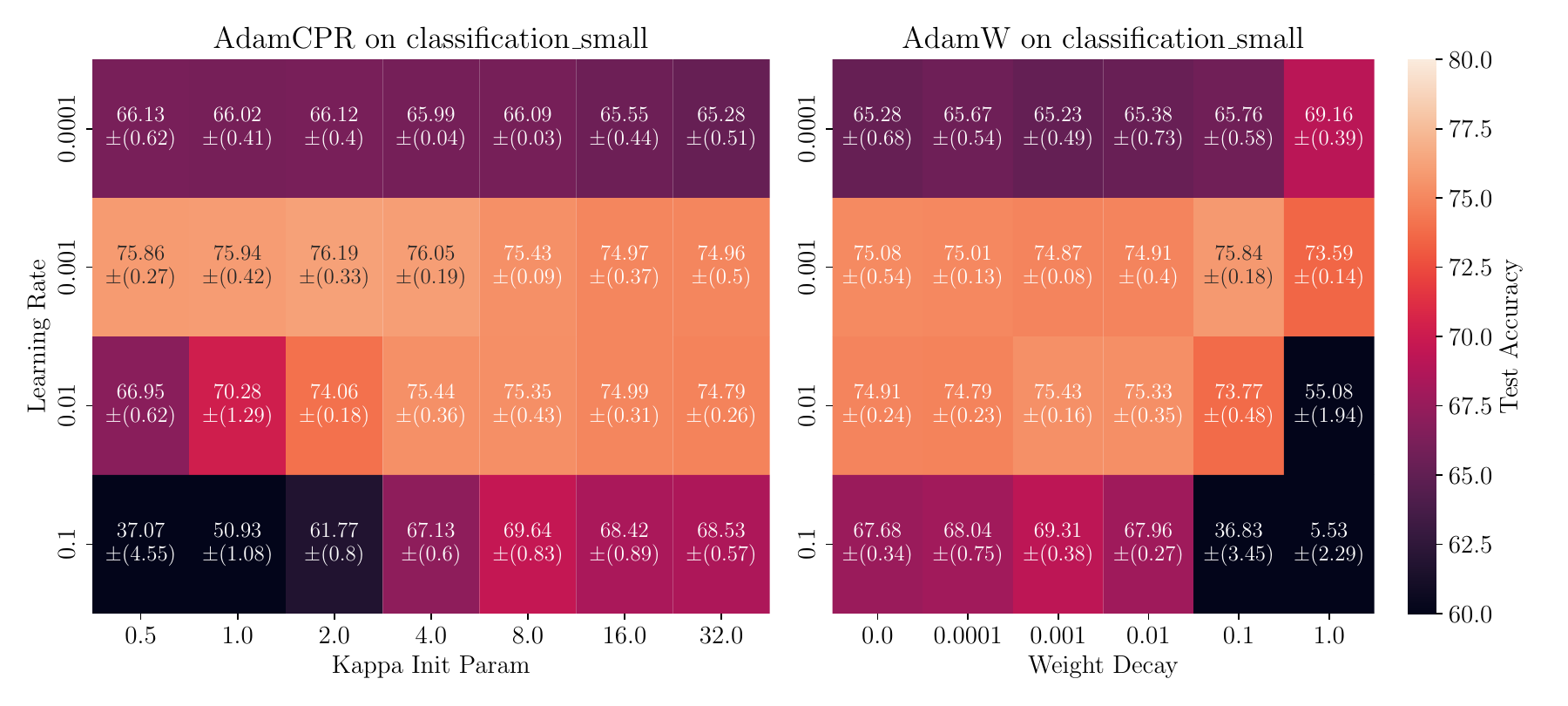}
	\caption{Hyperparameters matching those used in \parencite{adamcpr}, resulting accuracies closely match those reported in that work.}
	\label{fig:classification-small-paper-last}
\end{figure}